\documentclass[sigconf,natbib=true,anonymous=False]{acmart}
\usepackage{bm} 
\usepackage{csquotes}

\AtBeginDocument{%
  \providecommand\BibTeX{{%
    \normalfont B\kern-0.5em{\scshape i\kern-0.25em b}\kern-0.8em\TeX}}}

\copyrightyear{2023}
\acmYear{2023}
\setcopyright{acmlicensed}\acmConference[CIKM '23]{Proceedings of the 32nd ACM International Conference on Information and Knowledge Management}{October 21--25, 2023}{Birmingham, United Kingdom}
\acmBooktitle{Proceedings of the 32nd ACM International Conference on Information and Knowledge Management (CIKM '23), October 21--25, 2023, Birmingham, United Kingdom}
\acmPrice{15.00}
\acmDOI{10.1145/3583780.3615455}
\acmISBN{979-8-4007-0124-5/23/10}

\begin{document}

\title{Addressing Selection Bias in Computerized Adaptive Testing: \\ A User-Wise Aggregate Influence Function Approach} 





\author{Soonwoo Kwon}
\authornote{Both authors contributed equally to the paper}
\affiliation{Riiid AI Research%
    \city{Seoul}
    \country{Republic of Korea}
    }
\author{Sojung Kim}
\authornotemark[1]
\affiliation{Riiid AI Research%
    \city{Seoul}
    \country{Republic of Korea}
    }
\author{Seunghyun Lee}
\affiliation{Riiid AI Research%
    \city{Seoul}
    \country{Republic of Korea}
    }
\author{Jin-Young Kim}
\affiliation{Riiid AI Research%
    \city{Seoul}
    \country{Republic of Korea}
    }
\author{Suyeong An}
\affiliation{Riiid AI Research%
    \city{Seoul}
    \country{Republic of Korea}
    }
\author{Kyuseok Kim}
\affiliation{Riiid AI Research%
    \city{Seoul}
    \country{Republic of Korea}
    }


\renewcommand{\shortauthors}{Soonwoo Kwon et al.}


\begin{abstract}

Computerized Adaptive Testing (CAT) is a widely used, efficient test mode that adapts to the examinee's proficiency level in the test domain. 
CAT requires pre-trained item profiles, for CAT iteratively assesses the student real-time based on the registered items' profiles, and selects the next item to administer using candidate items' profiles. 
However, obtaining such item profiles is a costly process that involves gathering a large, dense item-response data, then training a diagnostic model on the collected data. 
In this paper, we explore the possibility of leveraging response data collected in the CAT service. 
We first show that this poses a unique challenge due to the inherent \textit{selection bias} introduced by CAT, i.e., more proficient students will receive harder questions. 
Indeed, when naïvely training the diagnostic model using CAT response data, we observe that item profiles deviate significantly from the ground-truth. 
To tackle the selection bias issue, we propose the user-wise aggregate influence function method.
Our intuition is to filter out users whose response data is heavily biased in an aggregate manner, as judged by how much perturbation the added data will introduce during parameter estimation. 
This way, we may enhance the performance of CAT while introducing minimal bias to the item profiles. 
We provide extensive experiments to demonstrate the superiority of our proposed method based on the three public datasets and one dataset that contains real-world CAT response data.


\end{abstract}

\ccsdesc[500]{Applied computing~E-learning}
\ccsdesc[500]{Social and professional topics~Student assessment}




\keywords{Computerized Adaptive Testing; Influence Function; Bias Reduction}



\maketitle

\section{Introduction}

Computerized Adaptive Testing (CAT) is of significant importance in both traditional assessment and digital learning~\cite{WK1984, ijcai2021p332, Bi2020quality, rcat, Settles20, zhuang2022fully}. CAT adaptively selects questions that align with the test taker's proficiency, minimizing redundancy and irrelevant questions while ensuring precise measurement~\cite{CAT2000}. As a result, CAT has been widely used for standard assessments such as GMAT, GRE, MCAT, and PISA. This efficiency benefits not only large-scale examinations but also extends to formative assessment practices. With the widespread adoption of digital learning systems, including remote learning in the COVID-19 era, CAT can facilitate frequent formative assessment by offering a daily personalized adaptive testing throughout the learning process~\cite{app10228196,YANG2022100104}.  

To present questions adaptively to the user's estimated skill level, CAT crucially depends on item (question\footnote{The terms \enquote{item} and \enquote{question} are used interchangeably to denote the individual questions of a given test.}) profiles. 
CAT assesses the user real-time based on the administered items' profiles, and selects the next question accordingly based on the unadministered items' profiles. 
Obtaining such item profiles is a costly procedure that involves collecting a full response dataset, i.e., for a larger number of test-takers, we must register every single question under consideration. 
This motivates us to consider a more data-efficient approach: starting with a smaller response dataset, and continuously updating the item profiles by leveraging data collected from the CAT service. 

\begin{figure}
    \centering    
    \includegraphics[width=\columnwidth]{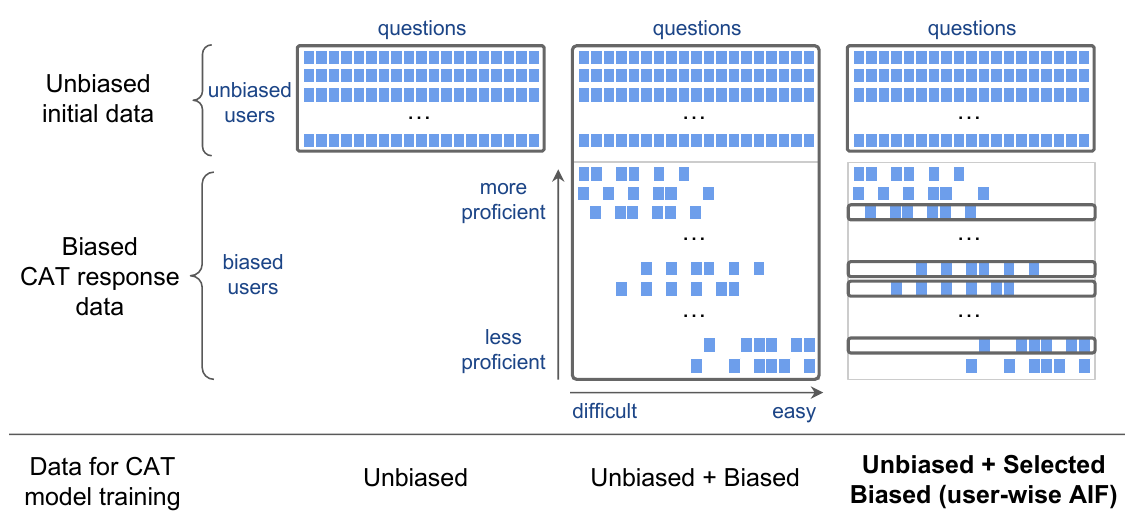}
    \vspace{-0.5cm}
    \caption{Illustration of CAT selection bias in user-item response patterns and our proposed approach}
    \label{fig:concept}
    \vspace{-0.5cm}
\end{figure}

However, we observe that leveraging CAT response data poses a challenge due to \textbf{\textit{selection bias}}. In CAT, this bias naturally arises because the difficulty of administered questions corresponds to the student's proficiency level. A similar form of selection bias is observed in domains like online recommendation systems, where user responses are shaped more by system algorithms than by genuine user preferences \cite{ovaisi2020correcting, wang2021combating}. We delve deeper into this aspect of CAT response data in Figure \ref{fig:concept} (middle).
In the figure, questions and users are arranged in order of difficulty and proficiency level, respectively. 
Note that CAT response data shows a distinct pattern where more proficient users tend to receive more difficult questions, while less proficient users receive easier questions. 
Such selection bias introduces distortions in item profiling; for instance, correct answers from advanced users for difficult questions may falsely inform the diagnostic model that these questions are not as difficult. 

To best leverage the CAT response data, we propose user-wise Aggregate Influence Function (AIF) method that selects minimally biased CAT response data before item profiling (see Figure~\ref{fig:concept} for an illustration). 
In particular, our method inspired by the Influence Function (IF) approach~\cite{KL2017, yu2020influence, chen2023survey, Silva2020UsingCI}, which identifies the contribution of each biased data point to the deviation of estimated item profile parameters. 
User-wise AIF, instead of focusing on IF of each response data point alone, considers the aggregate effect of a given user's entire response data, then filters out users whose responses are heavily biased. 

In our experiments, we show the effectiveness of User-wise AIF method based on three public datasets and one dataset that additionally contains real-world CAT response data. 
In all cases considered, our method is shown to outperform baseline methods in terms of assessment accuracy (as measured by AUC), while introducing minimal bias in item profiles (as measured by rank correlation between ground-truth and estimated difficulty levels). \footnote{The code and partial datasets are available at \url{https://github.com/riiid/UserAIF}.}

In summary, our contributions are:
\begin{itemize}
  \item Observation of selection bias in CAT response data: We identify the inherent selection bias that arises from the adaptive nature of CAT. Specifically, to the best of our knowledge, this is the first study to investigate selection bias in CAT response data using real-world CAT service data.
  \item User-wise Aggregate IF Method: We propose a novel debiasing method called the user-wise Aggregate IF approach. This simple method effectively mitigates the selection bias present in CAT response data, and overcomes the limitations of traditional IF approaches. 
  \item Experimental Results: Our experiments based on both simulated and real-world CAT response data demonstrate that incorporating CAT response data using AIF improves CAT performance, even with limited unbiased initial data. 
  \end{itemize}


The paper unfolds as follows: In Section \ref{sec:cat}, we address the selection bias in CAT with a real-world dataset. Our user-wise AIF approach is introduced in Section \ref{sec:aif}, followed by experimental results in Section \ref{sec:experiments}.

\section{CAT and Selection Bias}\label{sec:cat}
{\bf Computerized Adaptive Testing.}
CAT~\cite{CAT2000} consists of two key components: real-time user assessment and question selection. 
Firstly, a given user's proficiency level is estimated real-time based on the pre-trained item profiles. 
For dichotomous items, commonly used diagnostic models for item profiling include one- to four-parameter logistic Item Response Models (IRT)~\cite{reise2014handbook}.
The second key component of CAT is the question selection algorithm. 
This algorithm recommends the next item to be presented to the test taker, taking into account their proficiency level and the candidate items' profiles. 
Over the years, various item selection algorithms have been researched extensively~\cite{Weiss82, Lord80, WC11, Chang1996}. 
More recently, some studies have also explored training the question selection algorithm by formulating the CAT task as a bi-level optimization problem~\cite{ijcai2021p332} or as a reinforcement learning problem~\cite{zhuang2022fully}.

\noindent {\bf Preliminaries.}
As described above, CAT crucially depend on item profiles obtained via diagnostic modeling prior to CAT. We first describe how the item profiles are learned based on diagnostic modeling and the initially collected dense data. 

Suppose for a set of users $\mathcal{U}$ and a set of questions $\mathcal{Q}$, we have an item-response data $\{x_{ij}| i\in \mathcal{U}, j \in \mathcal{Q}\}$, where $x_{ij} = 1$ if the user $i$ answered question $j$ correctly, and $0$ otherwise. 
The goal of diagnostic modeling is to accurately predict the probability that a given user answers a given question correctly: 
\begin{equation*}
    p_{ij} = \mathsf{P}(x_{ij} = 1 | d_j, \theta_i),
\end{equation*}
where $\theta_i \in \mathbb{R}^{1 \times m}$ denotes the trainable $m$-dimensional user embedding (proficiency parameter) for user $i \in \mathcal{U}$, and $d_j \in \mathbb{R}^{1 \times n}$ denotes the $n$-dimensional item embedding for item $j \in \mathcal{Q}$. 

To train the model, the following individual loss is assigned to each data point given the respective parameters:
\begin{equation}\label{eq:loss}
    l(x_{ij} | d_j, \theta_i) =  - x_{ij} \log p_{ij} - (1- x_{ij}) \log (1 - p_{ij}).
\end{equation}
Finally, given the initially collected dense, unbiased dataset, we minimize the following loss function: 
\begin{equation}\label{eq:risk}
    L(\bm{x} | \bm{\beta}) = \frac{1}{\sum_{i \in \mathcal{U}_{unbiased}} |\mathcal{Q}_i|} \sum_{i \in \mathcal{U}_{unbiased}} \sum_{j \in \mathcal{Q}_i} l(x_{ij} | d_j, \theta_i), 
\end{equation}
where $\mathcal{U}_{unbiased}$ denotes the set of users in the dense, unbiased dataset, and $\bm{\beta} = \left[d_1, \cdots, d_{|\mathcal{Q}|}, \theta_1, \cdots, \theta_{|\mathcal{U}_{unbiased}|}\right]$ denotes all the parameters to be estimated for the unbiased training set $\bm{x}$ using the diagnostic model. 
Thus, we refer to $\mathcal{U}_{unbiased}$ as \textit{unbiased users}.


\noindent {\bf Selection Bias.}
Assuming that the item embeddings are prepared and an appropriate item selection algorithm has been established, we are ready to initiate the CAT service. 
Suppose that after launching the CAT service, we have collected response data from the serviced users, denoted as 
$$\bm{z} = \{\bm{z}_k\}_{k \in \mathcal{U}_{biased}} = \{ z_{kj}, j \in \mathcal{Q}_k \}_{k \in \mathcal{U}_{biased}}.$$
Similar to $\bm{x}$, $z_{kj}$ represents the correctness of a new user $k$'s response to an item $j \in \mathcal{Q}_k$. The set $\mathcal{U}_{biased}$ represents the group of users who have received CAT services, and we refer to them as \textit{biased users}. 
While the set of questions $\mathcal{Q}_i$ for an unbiased user $i$ is the same as $\mathcal{Q}$, $\mathcal{Q}_k$ is a personalized question set administered for each user $k$.

The personalized nature of CAT introduces bias into the CAT response dataset $\bm{z}$. 
For example, for the one-dimensional one-parameter IRT model, Fisher Information (FI) based item selection algorithm~\cite{Weiss82, Lord80} chooses an item that maximizes the FI 
\begin{equation*}
    I_{j}(\theta_k) = p_{kj}(1-p_{kj}).
\end{equation*}
This implies that the algorithm tends to prefer items with predicted correctness probabilities close to 0.5. As a result, a strong relationship between item parameters and user proficiency is explicitly introduced, which contributes to selection bias in the dataset $\bm{z}$.

\noindent {\bf Selection Bias in Real-World Dataset.}
We now demonstrate that selection bias in CAT indeed occurs in the real world, based on a CAT response dataset collected from R.test\footnote{\url{https://www.rtest.ai/}}, an AI-powered diagnostic test platform that assesses students' test readiness using CAT.
\begin{figure}
    \centering    
    \includegraphics[width=0.8\columnwidth]{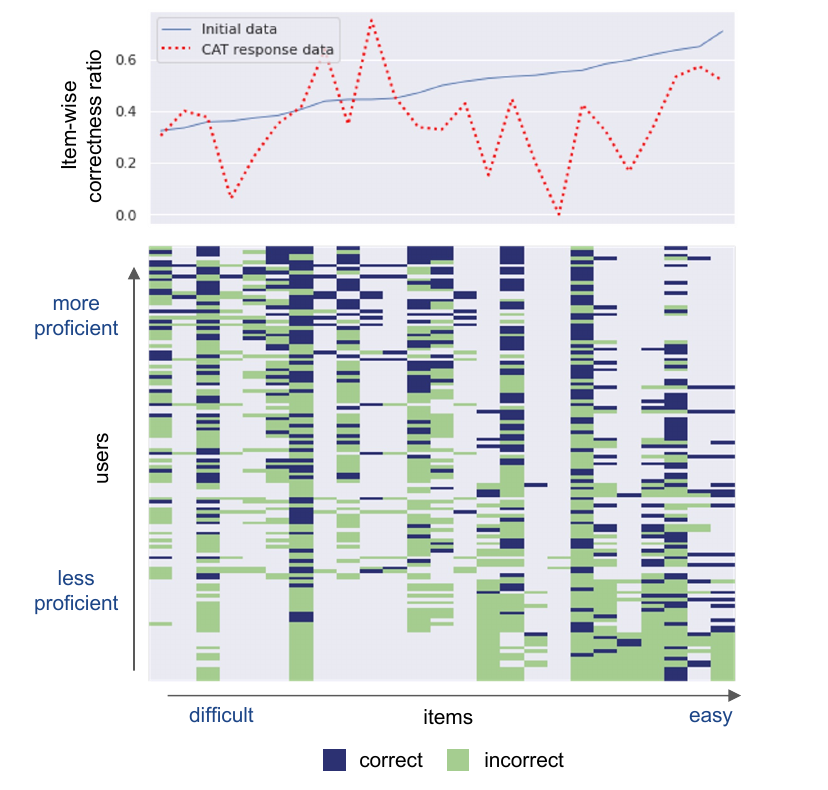}
    \vspace{-0.3cm}
    \caption{Visualization of user-item responses and correctness ratio in the SAT reading section on R.test platform.}
    \label{fig:rtest}
    \vspace{-0.5cm}
\end{figure}
The lower panel of Figure \ref{fig:rtest} presents the user-item responses for the SAT reading section, which consists of 52 questions. The CAT service dataset contains responses from 125 serviced users, where each user solved 10 questions. 
Similar to Figure \ref{fig:concept}, the items and users are ordered based on their correctness ratio. 
Due to some content constraints, the user-item pattern in Figure \ref{fig:rtest} appears slightly different from the one in Figure \ref{fig:concept}. 
Nevertheless, it still exhibits the tendency that the difficulty of the items received by users correlates with their ability levels.

The upper panel of Figure \ref{fig:rtest} further illustrates this phenomenon by displaying the correctness ratio for all items, using both unbiased response data and CAT service data. It is evident that the CAT service data deviates significantly from the original correctness ratio.
To further analyze the extent of this bias, we compute the rank correlation between the item parameters obtained from a large unbiased dataset and those obtained from the biased CAT service data. The observed rank correlation was 0.0881. 
The finding confirms the presence of bias in the CAT service data, highlighting the need for effective methods to mitigate this bias and enhance the accuracy of item parameter estimation in CAT.

\section{Aggregate Influence Function}\label{sec:aif}
In this section, we describe our proposed method, user-wise Aggregate Influence Function (AIF), designed to mitigate the selection bias in CAT response data. 
We first introduce the Influence Function (IF) approach. 
Next, we develop IF to be more suitable for CAT response data by considering the aggregate IF on a per-user basis. 

\subsection{Influence function}
In simple terms, IF measures the amount of perturbation introduced in the estimated parameters when a data point is added to the existing training dataset. 
More formally, recall that we have obtained the parameters $\hat{\bm{\beta}} \in \mathbb{R}^{1 \times \left(n|\mathcal{Q}|+m|\mathcal{U}_{unbiased}| \right)}$
from the unbiased data $\bm{x}$ by optimizing \eqref{eq:risk}, that is,  
\begin{equation*}
    \hat{\bm{\beta}} = \arg \min_{\bm{\beta}} L(\bm{x} | \bm{\beta}).
\end{equation*}
Suppose now a single new data point $z_{kj}$ from $\bm{z}$, a response to an item $j$ for a biased user $k \in \mathcal{U}_{biased}$, is added to the training set $\bm{x}$. 
We let $\theta_k^{biased}$ be the user embedding of the biased user $k$.

To study its impact in the original parameter $\hat{\bm{\beta}}$, consider the perturbed parameter estimator 
\begin{equation*}
    \hat{\bm{\beta}}_{\epsilon,z_{kj}} = \arg \min_{\bm{\beta}} \ (1-\epsilon)L(\bm{x} | \bm{\beta}) + \epsilon l(z_{kj} | d_j, \theta_k^{biased}).
\end{equation*}
Here, instead of considering a new parameter for the biased user, we use the fixed value $\theta_k^{biased}$ estimated from CAT. 
Note that this estimate is fairly accurate when the CAT’s test length is large enough. 
Also, this approach enables us to simplify the problem by only considering the original parameters $\bm{\beta}$. 

Using Hampel's IF~\cite{Hampel74}, we assess the effect on $\bm{\beta}$ estimate of disturbing empirical distribution on the unbiased set $\bm{x}$ by an infinitesimal amount of contamination at the new data point $z_{kj}$: 
\begin{equation}\label{eq:if_def}
    \mathsf{IF}_{\bm \beta}(z_{kj},\bm{x}) = \lim_{\epsilon \to 0} \frac{\hat{\bm{\beta}}_{\epsilon,z_{kj}} - \hat{\bm \beta}}{\epsilon}.
\end{equation}
This IF \eqref{eq:if_def} can be computed~\cite{Hampel74,KL2017} as 
\begin{equation}\label{eq:if}
    \mathsf{IF}_{\bm \beta}(z_{kj},\bm{x}) = - \bm{H}_{\bm{x}}^{-1} \left[\bigtriangledown_{\hat{\bm \beta}} l(z_{kj}) \right]^T,
\end{equation}
where $\left. \bm{H}_{\bm{x}} = \bigtriangledown_{\bm \beta}^2 L(\bm{x} | \bm{\beta}) \right|_{\bm{\beta} = \hat{\bm{ \beta}}}$ is the Hessian matrix and $\bigtriangledown_{\hat{\bm \beta}} l(z_{kj})$ is an abbreviation for $ \left. \bigtriangledown_{\bm \beta} l(z_{kj}| d_j, \theta_k^{biased}) \right|_{\bm \beta = \hat{\bm \beta}}$. 
For IRT models, both the gradient and the Hessian matrix can be defined and calculated analytically.
\footnote{The code implementing these formulas, along with its description, can be found at \url{https://github.com/riiid/UserAIF}.}


While the application of IF \eqref{eq:if} for debiasing problems vary\footnote{The baselines in Section \ref{sec:experiments} provide a list of the relevant applicable methods.}, the most widely adopted scheme was introduced in~\cite{KL2017}.
In their work, the authors introduce the use of IF on the loss computed at a new unbiased test point. This is derived using the chain rule and is expressed in our context as
\begin{equation*}
     \mathsf{IF}_{\mathsf{loss}}(z_{kj},y_{pq}) = - \bigtriangledown_{\hat{\bm \beta}} l(y_{pq}) \bm{H}_{\bm{x}}^{-1} \left[ \bigtriangledown_{\hat{\bm \beta}} l(z_{kj})\right]^T
\end{equation*}
Here, $y_{pq}$ represents the response of an unbiased user $p$ from $\mathcal{U}_{unbiased}^{\mathsf{val}}$, which is a set of unbiased validation users that does not overlap with $\mathcal{U}{unbiased}$. 
When considering an entire set $\bm{y}$ of $y_{pq}$, we abuse the notation to describe 
\begin{equation}\label{eq:if_loss_val}
    \mathsf{IF}_{\mathsf{loss}}(z_{kj},\bm{y}) = \sum_{p \in \mathcal{U}_{unbiased}^{\mathsf{val}}} \sum_{q \in \mathcal{Q}_p} \mathsf{IF}_{\mathsf{loss}}(z_{kj},y_{pq}).
\end{equation}

Both the IF \eqref{eq:if} on the estimated parameters and the IF \eqref{eq:if_loss_val} on the loss provide an approximation of the contribution of each biased data point to the deviation of the parameters and loss, respectively. However, when dealing with a large group of data points, it becomes necessary to consider the cross-effects within the group, as highlighted in a previous study~\cite{KATL2019}. Therefore, instead of focusing solely on the IFs of individual data points, we propose a data selection approach that takes into account groups of data, driven by the observed characteristics of selection bias in our study.


\subsection{User-wise aggregate influence function}

In what follows, we propose a data selection approach that addresses two key considerations. Firstly, we aim to minimize the influence of selected data points on the changes in the original parameters. Secondly, we endeavor to preserve the relative difficulty rankings of items within the selected group. To achieve this, our proposed method adopts a user-wise approach, incorporating the complete response data for each user to maintain the rank information within their dataset. Additionally, we leverage aggregate IF values for all parameters across the user's data points.

We first define the user-wise AIF of estimator $\hat{\bm \beta}$ for each biased user as follows. 
\begin{equation}\label{eq:aif}
    \mathsf{AIF}_{\bm \beta}(k,\bm{x}) =  \sum_{j \in \mathcal{Q}_k } \ \sum_{c} \left[ \mathsf{IF}_{\bm \beta}(z_{kj},\bm{x}) \right]_c 
\end{equation}
for user $k \in \mathcal{U}_{biased}$. 
Here, the index $c$ ranges from 1 to the total length of the vector $\bm{\beta}$ and $\left[ \mathsf{IF}_{\bm \beta}(z_{kj},\bm{x}) \right]_c$ represents the the $c$-th element of $\mathsf{IF}_{\bm \beta}(z_{kj},\bm{x})$.

Based on AIF, we propose the following selection method, denoted as "${\mathsf{User}}\mathsf{AIF}$":
\begin{itemize}
    \item[1.] Compute $\mathsf{AIF}_{\bm \beta}(k,\bm{x})$ for every biased user $k$.
    \item[2.] Choose a set of selected biased users $\mathcal{U}_{biased}^s$ whose $| \mathsf{AIF}_{\bm \beta}(k,\bm{x})|$ is less than some threshold $\eta$, that is,
    \begin{equation*}
        \mathcal{U}_{biased}^s = \left\{ k \in \mathcal{U}_{biased}: | \mathsf{AIF}_{\bm \beta}(k,\bm{x}) | \leq \eta \right\}.
    \end{equation*}
    \item[3.] Retrain model with $\bm{x} \cup \{ \bm{z}_k \}_{k \in \mathcal{U}_{biased}^s}$.
\end{itemize}
In the experiment, the threshold $\eta$ was chosen as the first quantile of the set $\{ | \mathsf{AIF}_{\bm \beta}(k,\bm{x}) | \}_{k \in \mathcal{U}_{biased}}$,
but the numerical result was robust to the choice. 
%

\section{Experiments}\label{sec:experiments}
In this section, we present the experimental results conducted on three public datasets and one real-world CAT response dataset.
We begin with the description of the baseline approaches that we compare our method against. 
Subsequently, we present a comparative analysis of our method using two public dense datasets. 
Furthermore, we explore the effectiveness of our method when applied to a recently developed neural model for CAT as well as the real-world CAT response data collected from R.test.

\subsection{Baselines}
In the following description of the baselines, we use the notations $\bm{x}$, $\bm{y}$, and $\bm{z}$ to represent the unbiased training set, validation set, and biased set, respectively.

\subsubsection{Retraining} The simplest baseline is to retrain the item response model with the changed dataset. We use three types of datasets: biased ($\bm{z}$), unbiased ($\bm{x}$), and both biased and unbiased ($\bm{z} \cup \bm{x}$). 

\subsubsection{IPS} Weighting by Inverse of Propensity Score (IPS) is one of the most popular method for debiasing in recommendation system. However, it is difficult to estimate the propensity score with sparse dataset, so we employ naive frequency based IPS method, denoted as IPS-NB and IPS-NB-IW, in~\cite{yu2020influence}.

\subsubsection{IF} We experiment on various baselines using IF. There are two types of IF based baselines, a weighting method and a selection from a biased set using IF.
\begin{itemize}
    \item IF4URec~\cite{yu2020influence}: It is the most relevant work that proposes an interaction-wise weighting method and demonstrates effectively debiases in recommendation. The suggested weight value for an interaction is computed as follows using \eqref{eq:if_loss_val}:
        \begin{equation*}
        \pi_{k,j}=\frac{1}{1+e^{\frac{\alpha \mathsf{IF}_{\mathsf{loss}}(z_{kj},\bm{y})}{\max_{k,j} \left(\left\{\mathsf{IF}_{\mathsf{loss}}(z_{kj},\bm{y}))\right\}\right)- \min_{k,j} \left(\left\{\mathsf{IF}_{\mathsf{loss}}(z_{kj},\bm{y}))\right\}\right)}}},
        \end{equation*}
     for some constant $\alpha$.
    
    \item $\mathsf{IF}_{\mathsf{loss}}$: 
    We select data $z_{kj}$ in $\bm{z}$ satisfying $\mathsf{IF}_{\mathsf{loss}}(z_{kj},\bm{y})) < 0$ where $\mathsf{IF}_{\mathsf{loss}}(z_{kj},\bm{y})$ is computed using equation \eqref{eq:if_loss_val}.
    The selected points are expected to minimize validation loss. We then train the final model using the unbiased set and the selected data, i.e. $\bm{x} \cup \{z_{kj} :  \mathsf{IF}_{\mathsf{loss}}(z_{kj},\bm{y})) < 0\}$. 
    
    \item $\mathsf{IF}_{\mathsf{param}}$: 
    We select data $z_{kj}$ in $\bm{z}$ satisfying 
    $\mathsf{IF}_{\bm \beta}(z_{kj},\bm{x})) < \eta$ 
    for some threshold $\eta$ using equation \eqref{eq:if}. 
    The selected interactions are expected to result in a perturbed parameter that is close to the original parameter trained on unbiased set.
    Data for retraining is then $\bm{x} \cup \{z_{kj} : \mathsf{IF}_{\bm \beta}(z_{kj},\bm{x}) < \eta \}$.

    \item ${\mathsf{Greedy}}\mathsf{AIF}$: 
    We sequentially find a group of data points that maximizes the aggregation effect in a greedy fashion. 
    From an initial point $z_{kj}^1 \in \bm{z}$, the $S$-th chosen data, $z_{pq}^{S}$, will be
    $\arg \min_{p, q} \left\{ \sum_{c} [\mathsf{IF}_{\bm \beta}(z_{pq},\bm{x})]_c + {\sum_{s=1}^{S-1} \sum_{c} [ \mathsf{IF}_{\bm \beta}(z_{p_s q_s}^s,\bm{x})]_c} \right\}.$ \\
\end{itemize}

\subsection{Main comparative experiments}

Throughout this section, we use one-dimensional two-parameter IRT model with Kullback-Leibler Information (KLI) based item selection method~\cite{Chang1996}. 
This is widely recognized as the most popular approach in CAT literature~\cite{WCB11,Yao13} and has been shown to deliver the best performance for the datasets under consideration. 

\subsubsection{Datasets}\label{sec:datasets}
In the absence of a public dataset with CAT bias, we make biased data by simulating CAT on two real-world dense datasets: ENEM and NIPS-EDU. 
ENEM is from 2019 Exame Nacional do Ensino Medio examination.
We use the densified NIPS-EDU~\cite{wang2020instructions}, released in NeurIPS 2020 Education Challenge from Eedi.
For each dataset, unbiased users are splitted into training and validation users, $\mathcal{U}_{unbiased}$ and $\mathcal{U}_{unbiased}^{\mathsf{val}}$, respectively.
We obtain the ENEM and NIPS-EDU CAT datasets, which are the biased sets that contain CAT bias generated from KLI method. 
Considering the total number of questions, $|\mathcal{Q}|$, we select 30 items for ENEM and 40 items for NIPS-EDU for each biased user in $\mathcal{U}_{biased}$.
Statistics for each dataset are described in Table \ref{table:data_statistics}. Note that we mimic a realistic scenario where the number of unbiased users is limited.

\begin{table}[h]
\vspace{-0.2cm}
  \caption{Statistics of real-world datasets}
  \vspace{-0.2cm}
  \label{table:data_statistics}
  \resizebox{\columnwidth}{!}{
      \begin{tabular}{lccccc}
        \toprule
        Datasets & $|\mathcal{Q}|$ & $|\mathcal{U}_{unbiased}|$ & $|\mathcal{U}_{unbiased}^{\mathsf{val}}|$ & $|\mathcal{U}_{biased}|$ & \# test users  \\
        \midrule
        ENEM & 185 & 40 & 10 & 1,000 & 2,000  \\
        \midrule 
        NIPS-EDU & 200 & 40 & 10 & 638 & 200 \\
        \midrule 
        Assist2009 & 26,700 & 266 & 66 & 334 & 166 \\
        \bottomrule
      \end{tabular}
  }
\vspace{-0.3cm}
\end{table}

\begin{table*}[h]
  \centering
  \caption{
    Performances of the methods in two simulated real-world data sets.
  }
  \vspace{-0.2cm}
  \resizebox{\textwidth}{!}
  {
 \begin{tabular}{c|cccccc|cccccc}

\toprule
\multicolumn{1}{c|}{Dataset} & \multicolumn{6}{c|}{ENEM} & \multicolumn{6}{c}{NIPS-EDU} \\
\hline
\multicolumn{1}{c|}{Metric} & 
\multicolumn{2}{c|}{CAT AUC-Pool} & \multicolumn{2}{c|}{CAT AUC-Eval} & \multicolumn{1}{c}{Random AUC} & \multicolumn{1}{c|}{Rank Corr.} & \multicolumn{2}{c|}{CAT AUC-Pool} & \multicolumn{2}{c|}{CAT AUC-Eval} &  \multicolumn{1}{c}{Random AUC} & \multicolumn{1}{c}{Rank Corr.} \\
\hline
Step & 10 & \multicolumn{1}{c|}{20} & 10 & \multicolumn{1}{c|}{20} & $-$ & $-$ & 10 & \multicolumn{1}{c|}{20} & 10 & \multicolumn{1}{c|}{20} & $-$ & $-$ \\ 
\midrule
biased  &$63.03_{\pm 0.46}$ &\multicolumn{1}{c|}{$63.61_{\pm 0.50}$} &$64.06_{\pm 1.16}$ &\multicolumn{1}{c|}{$64.86_{\pm 1.00}$} & $63.20_{\pm 0.30}$ & $0.2924_{\pm 0.03}$ 
&$71.72_{\pm 0.75}$ &\multicolumn{1}{c|}{$73.00_{\pm 0.72}$} &$71.85_{\pm 1.22}$ &\multicolumn{1}{c|}{$73.14_{\pm 1.11}$} & $74.27_{\pm 5.25}$ & $0.3873_{\pm 0.05}$ \\

unbiased &$68.57_{\pm 0.67}$ &\multicolumn{1}{c|}{$69.64_{\pm 0.77}$} &$68.47_{\pm 0.86}$ &\multicolumn{1}{c|}{$69.56_{\pm 0.69}$} & $70.37_{\pm 0.72}$ & $0.8298_{\pm 0.03}$ 
&$76.37_{\pm 0.39}$ &\multicolumn{1}{c|}{$78.12_{\pm 0.23}$} &$76.28_{\pm 1.27}$ &\multicolumn{1}{c|}{$77.78_{\pm 1.20}$} & $78.49_{\pm 0.44}$ & $0.8644_{\pm 0.24}$ \\

biased $\cup$ unbiased &$68.31_{\pm 0.92}$ &\multicolumn{1}{c|}{$69.12_{\pm 0.91}$} &$68.59_{\pm 1.35}$ &\multicolumn{1}{c|}{$69.36_{\pm 1.34}$} & $70.04_{\pm 1.17}$ & $0.8134_{\pm 0.03}$ 
&$76.37_{\pm 0.47}$ &\multicolumn{1}{c|}{$78.05_{\pm 0.27}$} &$76.33_{\pm 1.20}$ &\multicolumn{1}{c|}{$77.74_{\pm 1.34}$} & $78.61_{\pm 3.19}$ & $0.8988_{\pm 0.03}$\\
 \hline
 
IPS-NB &$68.92_{\pm 1.29}$ &\multicolumn{1}{c|}{$69.53_{\pm 1.27}$} &$69.30_{\pm 1.00}$ &\multicolumn{1}{c|}{$70.06_{\pm 0.92}$} & $70.91_{\pm 1.11}$ & $0.8579_{\pm 0.03}$ 
&$76.31_{\pm 0.14}$ &\multicolumn{1}{c|}{$77.85_{\pm 0.12}$} &$76.32_{\pm 1.35}$ &\multicolumn{1}{c|}{$77.66_{\pm 1.34}$} & $78.65_{\pm 0.26}$ & $0.8992_{\pm 0.02}$ \\

IPS-NB-IW &$67.38_{\pm 2.28}$ &\multicolumn{1}{c|}{$68.75_{\pm 2.09}$} &$67.75_{\pm 1.32}$ &\multicolumn{1}{c|}{$68.78_{\pm 1.18}$} & $69.04_{\pm 1.85}$ & $0.7808_{\pm 0.09}$ 
&$75.79_{\pm 0.40}$ &\multicolumn{1}{c|}{$77.12_{\pm 0.48}$} &$75.73_{\pm 1.20}$ &\multicolumn{1}{c|}{$76.94_{\pm 1.29}$} & $77.81_{\pm 0.31}$ & $0.7856_{\pm 0.09}$ \\
\hline

IF4URec &$68.30_{\pm 0.92}$ &\multicolumn{1}{c|}{$69.22_{\pm 0.80}$} &$68.56_{\pm 1.55}$ & \multicolumn{1}{c|}{$69.57_{\pm 1.29}$} & $70.24_{\pm 1.08}$ & $0.8232_{\pm 0.02}$ 
&$76.39_{\pm 0.35}$ & \multicolumn{1}{c|}{$78.10_{\pm 0.28}$} &$76.38_{\pm 0.91}$ & \multicolumn{1}{c|}{$77.80_{\pm 1.19}$} & $78.64_{\pm 0.26}$ & $0.9002_{\pm 0.03}$ \\

$\mathsf{IF}_{\mathsf{loss}}$ & $66.44_{\pm 1.44}$ &\multicolumn{1}{c|}{$67.17_{\pm 0.85}$} & $67.10_{\pm 2.35}$ &\multicolumn{1}{c|}{$67.79_{\pm 2.05}$} & $67.96_{\pm 1.06}$ & $0.7088_{\pm 0.07}$ 
& $71.96_{\pm 1.06}$ & \multicolumn{1}{c|}{$72.92_{\pm 0.54}$} & $72.57_{\pm 1.16}$ & \multicolumn{1}{c|}{$73.49_{\pm 1.39}$} & $74.44_{\pm 9.60}$ & $0.5391_{\pm 0.06}$\\

$\mathsf{IF}_{\mathsf{param}}$ &$68.63_{\pm 0.53}$ & \multicolumn{1}{c|}{$69.84_{\pm 0.58}$} &$68.68_{\pm 0.93}$ & \multicolumn{1}{c|}{$69.64_{\pm 0.89}$} & $70.35_{\pm 0.46}$ & $0.8101_{\pm 0.03}$ 
& $75.25_{\pm 0.60}$ & \multicolumn{1}{c|}{$77.41_{\pm 0.48}$} & $75.35_{\pm 0.53}$ & \multicolumn{1}{c|}{$77.11_{\pm 0.80}$} & $78.46_{\pm 0.30}$ & $0.8731_{\pm 0.02}$ \\

$\mathsf{Greedy}\mathsf{AIF}$ &$68.12_{\pm 0.69}$ &\multicolumn{1}{c|}{$69.51_{\pm 0.57}$} &$68.21_{\pm 0.91}$ &\multicolumn{1}{c|}{$69.26_{\pm 0.77}$} & $69.99_{\pm 0.44}$ & $0.7869_{\pm 0.02}$ 
& $75.79_{\pm 0.65}$ &\multicolumn{1}{c|}{$77.62_{\pm 0.63}$} & $75.80_{\pm 0.47}$ &\multicolumn{1}{c|}{$77.22_{\pm 0.76}$} & $78.55_{\pm 0.32}$ & $0.8811_{\pm 0.02}$ \\

${\mathsf{User}}\mathsf{AIF}$ & $\textbf{69.18}_{\pm 0.54}$ & \multicolumn{1}{c|}{$\textbf{70.13}_{\pm 0.60}$} & $\textbf{69.30}_{\pm 0.83}$ & \multicolumn{1}{c|}{$\textbf{70.15}_{\pm 0.85}$} & $\textbf{71.00}_{\pm 0.45}$ & $\textbf{0.8600}_{\pm 0.03}$ 
& $\textbf{76.97}_{\pm 0.24}$ &\multicolumn{1}{c|}{ $\textbf{78.33}_{\pm 0.22}$} & $\textbf{77.12}_{\pm 1.46}$ &\multicolumn{1}{c|}{$\textbf{77.96}_{\pm 1.44}$} & $\textbf{78.87}_{\pm 0.33}$ & $\textbf{0.9030}_{\pm 0.00}$ \\

\bottomrule
\end{tabular}%
\label{table:result_main}
\normalsize}
\vspace{-0.3cm}
\end{table*}

\subsubsection{Experimental Settings} 

For evaluation metrics, we use Area Under the Receiving Operator Curve (AUC, represented in \%), Rank Correlation (Rank Corr.), and Accuracy(ACC, represented in \%). 
The $\pm$ values presented in the tables represent the standard deviation derived from five experiments, illustrating the consistency of our results across different runs. 
First, we inspect CAT AUC to evaluate the performance of CAT. To observe CAT AUC, we split the items into pool-items and eval-items. We perform CAT with pool-items, and evaluate the performance on eval-items. 
It is also important to validate our method on the quality of item representation. Thus, we inspect two additional metrics, Random AUC and Rank Corr. To observe Random AUC, we freeze the trained item representations and randomly provide 80\% of interactions for each user, which are used for updating user representation. Then, we evaluate AUC on the remaining 20\% of interactions. We also observe Rank Corr. of difficulty parameters between the retrained item representations and the item representations obtained from large unbiased data, which we consider as ground truth.

For early stopping, we use the validation method specialized to the setting of CAT. The goal of CAT is to assess the new user's ability using the trained item representation. Thus, we perform validation on $\bm{y}$. For each validation step, we freeze the item representations obtained from training and provide 80\% of interactions of $\bm{y}$. We compute the validation loss based on the prediction on the remaining 20\% of interactions. 

\subsubsection{Results}
Firstly, we point out the impact of selection bias in the considered CAT datasets. 
Table \ref{table:result_main} describes the evaluation metrics for baselines and our proposed method. Notably, all the metrics for retraining IRT with the biased set show significant underperformance for both datasets. The low rank correlations of the difficulty parameters highlight the distortion of ranks in item difficulty due to CAT bias, as previously mentioned. 

Secondly, ${\mathsf{User}}\mathsf{AIF}$ outperforms to other baselines in terms of evaluating both the quality of item representation and CAT performance in Table \ref{table:result_main}, by considering the aggregate effect of IF and user-wise selection that preserves the rank information received by each user. 

The IPS-based methods, the second part in row, can be promising. However, they tend to be less precise under a data-scarce setting like NIPS-EDU because frequency of interactions is used to calculate IPS. In NIPS-EDU, the IPS-based methods exhibit underperformance in CAT AUC compared to retraining IRT baselines.

The IF-based baseline methods, which consider IF on the individual data points, do not provide a complete solution. IF4URec and $\mathsf{IF}_{\mathsf{loss}}$ rely on the additional unbiased validation set, which can be a disadvantage due to its small size. We also observed that ${\mathsf{Greedy}}\mathsf{AIF}$, which considers the aggregation effect of $\mathsf{IF}_{\mathsf{param}}$ among the selected data points, led to an overall relative underperformance and a decrease in Rank Corr. for the ENEM dataset. 

\begin{figure}[h]
    \centering
    \includegraphics[width=\columnwidth]{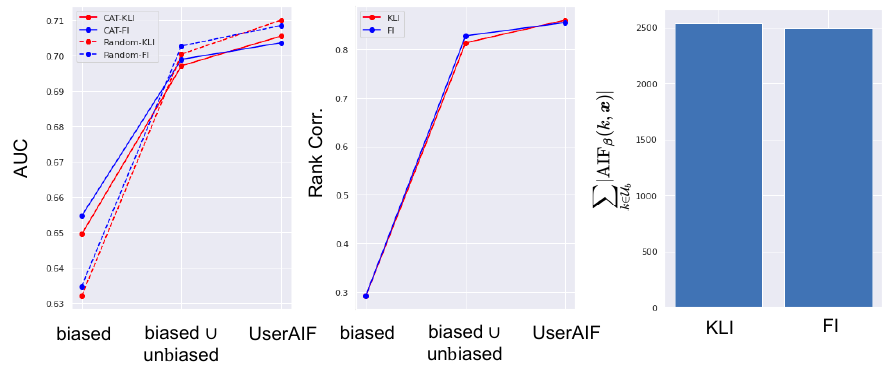}
    \vspace{-0.5cm}
    \caption{Ablation of item selection algorithms
    }
    \label{fig:method}
    \vspace{-0.3cm}
\end{figure}

Thirdly, we examine how the performance of ${\mathsf{User}}\mathsf{AIF}$ varies based on the degree of the selection bias induced by different item selection algorithms. Figure \ref{fig:method} presents the results for two biased sets generated by FI and KLI methods for the ENEM dataset. 
FI item selection algorithm~\cite{Weiss82, Lord80} is adopted for the two-parameter IRT.
The left panels show the AUC and Rank Corr., while the right panel displays the sum of the absolute values of AIF \eqref{eq:aif} for biased users.
We observe that the biased set generated by KLI exhibits poorer performance metrics and a larger sum of $| \mathsf{AIF}_{\bm \beta}(k,\bm{x}) |$, indicating that the KLI method introduces a stronger bias.
${\mathsf{User}}\mathsf{AIF}$ demonstrates greater improvement in all metrics for the KLI-biased set. We anticipate that our method will perform well for other item selection algorithms, as evidenced by both results for FI and KLI. 

\subsection{Applicability to neural model}
Although IRT-based response models and information-based item selection methods are still widely used for CAT, a few deep neural network-based methods have recently been introduced and have shown outstanding performance~\cite{ijcai2021p332, zhuang2022fully}. BOBCAT ~\cite{ijcai2021p332}, one of the SOTA methods, utilizes bi-level optimization to train an diagnostic model and an item selection model simultaneously. In particular, we use the BiNN-Approx model, which shows the best performance in the original paper. 

{\bf Dataset.} For our experiment, we use Assist2009 dataset that is one of the benchmark dataset used in the original paper. 
The Assist2009 data was gathered during the school year 2009-2010 from a web-based math tutoring system for 7th to 12th grade students. 
In contrast to the previous datasets, the Assist2009 dataset exhibits sparsity. In fact, the Assist2009 dataset has a filling rate of only 0.0128.
As a result, when conducting CAT, we can only select items that have existing interactions in the dataset. This constraint limits the available choices during the CAT procedure.
Given the requirement of deep models for sufficient training data, we incorporate a larger number of unbiased users in the Assist2009 dataset compared to the previous experiments. More detailed statistics about the Assist2009 dataset are presented in Table \ref{table:data_statistics}.
Using the trained BiNN-Approx model, we generate simulated biased CAT response data by administering a total of 30 questions in the CAT procedure.

{\bf Model.} 
The BiNN-Approx model takes a neural network as the response diagnostic model with the global response model parameters and the learned item selection algorithm with the approximate gradient on the question selection parameters. To apply our proposed method, we use the global response model parameters for ${\bm \beta}$ in the AIF computation \eqref{eq:aif}. Correspondingly, the dimension of ${\bm \beta}$ in the equation \eqref{eq:aif} is reduced to the product of the number of global response model parameters and the dimension of the parameters.

\begin{figure}[h!]
    \centering    
    \includegraphics[width=0.9\columnwidth]{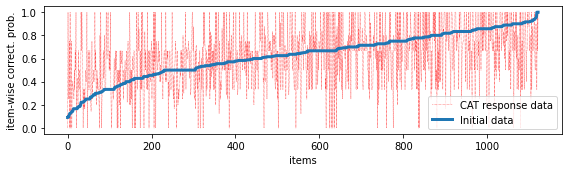}
    \vspace{-0.5cm}
    \caption{Visualization of correctness ratio in Assist2009.}
    \label{fig:catbias_bobcat}
    \vspace{-0.2cm}
\end{figure}

{\bf Results.} We begin by investigating whether selection bias occurs when using the BiNN-Approx model. 
Previous work \cite{ijcai2021p332} has mentioned the tendency of the learned item selection model to provide a less diverse set of items. However, we still observe the presence of selection bias in the generated CAT data. Figure \ref{fig:catbias_bobcat} illustrates this by comparing the correctness ratio of items that were administered more than 75\% in both the unbiased data and CAT response data. The biased CAT data (shown in red) exhibits fluctuations in the item correctness ratio, in contrast to the monotonic order observed in the unbiased data, similar to Figure \ref{fig:rtest}. Furthermore, Table \ref{table:result_bobcat} demonstrates a drop in CAT AUC when the biased data is added into the unbiased data.

Secondly, we show the performance of our method in the BiNN-Approx model at step 20 in Table \ref{table:result_bobcat}. 
Despite having a larger amount of unbiased data compared to the previous experiments, the ${\mathsf{User}}\mathsf{AIF}$ method demonstrates improvements in both AUC and ACC using only the unbiased dataset.
Moreover, when compared to IPS-NB, which was the strongest debiasing baseline in our previous experiments, the proposed method continues to outperform it.

\begin{table}[h]
  \caption{
    Performances of the methods used with BOBCAT
  }
  \resizebox{0.7\columnwidth}{!}
     {
     \begin{tabular}{c|cc}
    
    \toprule
    \multicolumn{1}{c|}{Dataset} & \multicolumn{2}{c}{Assist2009} \\
    \hline
    \multicolumn{1}{c|}{Metric} & \multicolumn{1}{c}{CAT AUC-Eval} & \multicolumn{1}{c}{CAT ACC-Eval} \\
    \midrule
    
    unbiased &$70.98_{\pm 0.53}$ & $68.99_{\pm 0.28}$  \\
    biased $\cup$ unbiased &$70.93_{\pm0.68}$ &$69.22_{\pm 0.52}$\\
    IPS-NB &$71.02_{\pm 0.96}$ &$68.65_{\pm 0.68}$ \\
    ${\mathsf{User}}\mathsf{AIF}$ & $\textbf{71.07}_{\pm 0.62}$ & $\textbf{69.26}_{\pm 0.41}$ \\
    \bottomrule
    \end{tabular}
    }
\label{table:result_bobcat}
\vspace{-0.3cm}
\end{table}

\subsection{Experiment with R.test CAT response data}
In this section, we verify the effectiveness of our proposed method by utilizing CAT service data collected by ourselves on the R.test platform, which is an AI-powered diagnostic test platform that assesses students' test readiness using CAT. 

The CAT model deployed in R.test was trained using a significant amount of initial response data to predict a student's SAT reading score after answering 10 out of 52 questions. 
Subsequently, we collected interactions from 125 users who utilized the service over a month, which served as the biased data for our analysis.
Additional constraints exist in the item selection algorithm, such as the requirement for preceding questions before certain question types like line evidence and sequential administration of questions related to the same reading passage. These constraints, inherent in real-world services, introduce additional bias sources.



Table \ref{table:result_sat} presents the evaluation metric results, including the Score Mean Absolute Error (MAE) for step $t = 3, 5$, which measures the difference between predicted and actual scores for the test users. The more precisely the CAT predicts a student's ability, the more accurately we can predict the student's score, i.e., the lower the MAE. 
The unbiased $\cup$ biased performance is inferior to the unbiased in multiple evaluation metrics. In contrast, both IPS-NB and our proposed method show improvements across all metrics. ${\mathsf{User}}\mathsf{AIF}$ performs exceptionally well in smaller steps, enhancing the user experience by accurately diagnosing users with fewer questions.

We acknowledge that the observed improvement in performance is relatively small, because the initial dataset was significantly larger than the biased set. However, we anticipate further performance gains as the service is used more extensively. Investigating and analyzing these potential improvements are left as future work.


\begin{table}[h]
  \caption{
    Performances of the methods in collected R.test data
  }
  \vspace{-0.2cm}
  \resizebox{\columnwidth}{!}
  {
 \begin{tabular}{c|cccccc}
\toprule
\multicolumn{1}{c|}{Dataset} & \multicolumn{6}{c}{SAT-reading} \\
\hline
\multicolumn{1}{c|}{Metric} & 
\multicolumn{2}{c|}{CAT AUC-Pool $\uparrow$} & \multicolumn{2}{c|}{MAE $\downarrow$} & \multicolumn{1}{c}{Random AUC $\uparrow$} & \multicolumn{1}{c}{Rank Corr. $\uparrow$}  \\
\hline
Step & 3 & \multicolumn{1}{c|}{5} & 3 & \multicolumn{1}{c|}{5} & $-$ & $-$ \\ 
\midrule

unbiased &$66.65_{\pm 0.70}$ &\multicolumn{1}{c|}{$69.10_{\pm 0.68}$} & $36.60_{\pm 0.79}$ & \multicolumn{1}{c|}{$29.43_{\pm 0.97}$} & $72.48_{\pm 1.05}$ &$0.9100_{\pm 0.005}$  \\

biased $\cup$ unbiased &$66.62_{\pm 0.96}$ &\multicolumn{1}{c|}{$69.04_{\pm 0.58}$} & $36.17_{\pm 1.65}$ &\multicolumn{1}{c|}{$29.01_{\pm 1.10}$}  &$72.40_{\pm 1.34}$ &$0.9101_{\pm 0.007}$ \\
 
IPS-NB &$67.06_{\pm 1.00}$ &\multicolumn{1}{c|}{$69.23_{\pm 0.57}$} & $35.77_{\pm 2.02}$ 
 &\multicolumn{1}{c|}{$\textbf{28.35}_{\pm 0.65}$} &$\textbf{72.51}_{\pm 1.07}$ & $0.9115_{\pm 0.008}$ \\

${\mathsf{User}}\mathsf{AIF}$ & $\textbf{67.31}_{\pm 0.75}$ & \multicolumn{1}{c|}{$\textbf{69.28}_{\pm 0.71}$} & $\textbf{35.63}_{\pm 0.96}$ &\multicolumn{1}{c|}{$28.66_{\pm 0.89}$} & $\textbf{72.51}_{\pm 1.08}$ & $\textbf{0.9119}_{\pm 0.008}$ \\

\bottomrule
\end{tabular}%
\label{table:result_sat}
}
\vspace{-0.3cm}
\end{table}

\section{Conclusion}\label{sec:conclusion}

We have observed the presence of selection bias in CAT response data, emerging from the adaptive nature of CAT. 
To mitigate this bias, we introduced the user-wise AIF method, which demonstrated superior performance compared to baseline approaches while introducing minimal bias to the item profiles. 
The method consistently outperformed across different settings, including variations in initial data, diagnostic model, and selection algorithm, using both simulated and real-world CAT response data. Future work includes the design of an adaptive data selection policy and exploring the application of second-order IF to examine group effects. This research highlights the feasibility of a data-efficient approach in launching CAT services, reducing data collection costs and time by leveraging the response data from the CAT service to continuously update CAT performance.

\appendix


\bibliographystyle{ACM-Reference-Format}
\balance
\bibliography{influence}


\begin{thebibliography}{25}


\ifx \showCODEN    \undefined \def \showCODEN     #1{\unskip}     \fi
\ifx \showDOI      \undefined \def \showDOI       #1{#1}\fi
\ifx \showISBNx    \undefined \def \showISBNx     #1{\unskip}     \fi
\ifx \showISBNxiii \undefined \def \showISBNxiii  #1{\unskip}     \fi
\ifx \showISSN     \undefined \def \showISSN      #1{\unskip}     \fi
\ifx \showLCCN     \undefined \def \showLCCN      #1{\unskip}     \fi
\ifx \shownote     \undefined \def \shownote      #1{#1}          \fi
\ifx \showarticletitle \undefined \def \showarticletitle #1{#1}   \fi
\ifx \showURL      \undefined \def \showURL       {\relax}        \fi
\providecommand\bibfield[2]{#2}
\providecommand\bibinfo[2]{#2}
\providecommand\natexlab[1]{#1}
\providecommand\showeprint[2][]{arXiv:#2}

\bibitem[Bi et~al\mbox{.}(2020)]%
        {Bi2020quality}
\bibfield{author}{\bibinfo{person}{H. Bi}, \bibinfo{person}{H. Ma},
  \bibinfo{person}{Z. Huang}, \bibinfo{person}{Y. Yin}, \bibinfo{person}{Q.
  Liu}, \bibinfo{person}{E. Chen}, \bibinfo{person}{Y. Su}, {and}
  \bibinfo{person}{S. Wang}.} \bibinfo{year}{2020}\natexlab{}.
\newblock \showarticletitle{Quality Meets Diversity: A Model-Agnostic Framework
  for Computerized Adaptive Testing}. In \bibinfo{booktitle}{\emph{2020 IEEE
  International Conference on Data Mining (ICDM)}}. \bibinfo{publisher}{IEEE
  Computer Society}, \bibinfo{address}{Los Alamitos, CA, USA},
  \bibinfo{pages}{42--51}.
\newblock


\bibitem[Chang and Ying(1996)]%
        {Chang1996}
\bibfield{author}{\bibinfo{person}{Hua-Hua Chang} {and}
  \bibinfo{person}{Zhiliang Ying}.} \bibinfo{year}{1996}\natexlab{}.
\newblock \showarticletitle{A Global Information Approach to Computerized
  Adaptive Testing}.
\newblock \bibinfo{journal}{\emph{Applied Psychological Measurement}}
  \bibinfo{volume}{20}, \bibinfo{number}{3} (\bibinfo{year}{1996}),
  \bibinfo{pages}{213--229}.
\newblock


\bibitem[Chen et~al\mbox{.}(2023)]%
        {chen2023survey}
\bibfield{author}{\bibinfo{person}{Jiawei Chen}, \bibinfo{person}{Hande Dong},
  \bibinfo{person}{Xiang Wang}, \bibinfo{person}{Fuli Feng},
  \bibinfo{person}{Meng Wang}, {and} \bibinfo{person}{Xiangnan He}.}
  \bibinfo{year}{2023}\natexlab{}.
\newblock \showarticletitle{Bias and Debias in Recommender System: A Survey and
  Future Directions}.
\newblock \bibinfo{journal}{\emph{ACM Trans. Inf. Syst.}} \bibinfo{volume}{41},
  \bibinfo{number}{3}, Article \bibinfo{articleno}{67} (\bibinfo{date}{feb}
  \bibinfo{year}{2023}).
\newblock
\showISSN{1046-8188}


\bibitem[Choi and McClenen(2020)]%
        {app10228196}
\bibfield{author}{\bibinfo{person}{Younyoung Choi} {and} \bibinfo{person}{Cayce
  McClenen}.} \bibinfo{year}{2020}\natexlab{}.
\newblock \showarticletitle{Development of Adaptive Formative Assessment System
  Using Computerized Adaptive Testing and Dynamic Bayesian Networks}.
\newblock \bibinfo{journal}{\emph{Applied Sciences}} \bibinfo{volume}{10},
  \bibinfo{number}{22} (\bibinfo{year}{2020}).
\newblock


\bibitem[Ghosh and Lan(2021)]%
        {ijcai2021p332}
\bibfield{author}{\bibinfo{person}{Aritra Ghosh} {and} \bibinfo{person}{Andrew
  Lan}.} \bibinfo{year}{2021}\natexlab{}.
\newblock \showarticletitle{BOBCAT: Bilevel Optimization-Based Computerized
  Adaptive Testing}. In \bibinfo{booktitle}{\emph{Proceedings of the Thirtieth
  International Joint Conference on Artificial Intelligence}}
  \emph{(\bibinfo{series}{{IJCAI'21}})},
  \bibfield{editor}{\bibinfo{person}{Zhi-Hua Zhou}} (Ed.).
  \bibinfo{publisher}{International Joint Conferences on Artificial
  Intelligence Organization}, \bibinfo{pages}{2410--2417}.
\newblock


\bibitem[Hampel(1974)]%
        {Hampel74}
\bibfield{author}{\bibinfo{person}{Frank~R. Hampel}.}
  \bibinfo{year}{1974}\natexlab{}.
\newblock \showarticletitle{The Influence Curve and its Role in Robust
  Estimation}.
\newblock \bibinfo{journal}{\emph{J. Amer. Statist. Assoc.}}
  \bibinfo{volume}{69}, \bibinfo{number}{346} (\bibinfo{year}{1974}),
  \bibinfo{pages}{383--393}.
\newblock


\bibitem[Koh and Liang(2017)]%
        {KL2017}
\bibfield{author}{\bibinfo{person}{Pang~Wei Koh} {and} \bibinfo{person}{Percy
  Liang}.} \bibinfo{year}{2017}\natexlab{}.
\newblock \showarticletitle{Understanding Black-Box Predictions via Influence
  Functions}. In \bibinfo{booktitle}{\emph{Proceedings of the 34th
  International Conference on Machine Learning}}
  \emph{(\bibinfo{series}{ICML'17})}. Article \bibinfo{articleno}{70},
  \bibinfo{numpages}{1885--1894}~pages.
\newblock


\bibitem[Koh et~al\mbox{.}(2019)]%
        {KATL2019}
\bibfield{author}{\bibinfo{person}{Pang Wei~W Koh}, \bibinfo{person}{Kai-Siang
  Ang}, \bibinfo{person}{Hubert Teo}, {and} \bibinfo{person}{Percy~S Liang}.}
  \bibinfo{year}{2019}\natexlab{}.
\newblock \showarticletitle{On the Accuracy of Influence Functions for
  Measuring Group Effects}. In \bibinfo{booktitle}{\emph{Advances in Neural
  Information Processing Systems}},
  \bibfield{editor}{\bibinfo{person}{H.~Wallach},
  \bibinfo{person}{H.~Larochelle}, \bibinfo{person}{A.~Beygelzimer},
  \bibinfo{person}{F.~d\textquotesingle Alch\'{e}-Buc},
  \bibinfo{person}{E.~Fox}, {and} \bibinfo{person}{R.~Garnett}} (Eds.),
  Vol.~\bibinfo{volume}{32}. \bibinfo{publisher}{Curran Associates, Inc.}
\newblock


\bibitem[Lord(1980)]%
        {Lord80}
\bibfield{author}{\bibinfo{person}{F.M. Lord}.}
  \bibinfo{year}{1980}\natexlab{}.
\newblock \bibinfo{booktitle}{\emph{Applications of Item Response Theory To
  Practical Testing Problems} (\bibinfo{edition}{1st.} ed.)}.
\newblock \bibinfo{publisher}{Routledge}.
\newblock


\bibitem[Ovaisi et~al\mbox{.}(2020)]%
        {ovaisi2020correcting}
\bibfield{author}{\bibinfo{person}{Zohreh Ovaisi}, \bibinfo{person}{Ragib
  Ahsan}, \bibinfo{person}{Yifan Zhang}, \bibinfo{person}{Kathryn Vasilaky},
  {and} \bibinfo{person}{Elena Zheleva}.} \bibinfo{year}{2020}\natexlab{}.
\newblock \showarticletitle{Correcting for Selection Bias in Learning-to-rank
  Systems}. In \bibinfo{booktitle}{\emph{Proceedings of The Web Conference
  2020}} \emph{(\bibinfo{series}{WWW '20})}. \bibinfo{publisher}{Association
  for Computing Machinery}, \bibinfo{address}{New York, NY, USA},
  \bibinfo{pages}{1863--1873}.
\newblock


\bibitem[Reise and Revicki(2014)]%
        {reise2014handbook}
\bibfield{author}{\bibinfo{person}{Steven~P Reise} {and}
  \bibinfo{person}{Dennis~A Revicki}.} \bibinfo{year}{2014}\natexlab{}.
\newblock \bibinfo{booktitle}{\emph{Handbook of item response theory
  modeling}}.
\newblock \bibinfo{publisher}{Taylor \& Francis New York}.
\newblock


\bibitem[Settles et~al\mbox{.}(2020)]%
        {Settles20}
\bibfield{author}{\bibinfo{person}{Burr Settles}, \bibinfo{person}{Geoffrey
  T.~LaFlair}, {and} \bibinfo{person}{Masato Hagiwara}.}
  \bibinfo{year}{2020}\natexlab{}.
\newblock \showarticletitle{Machine Learning–Driven Language Assessment}.
\newblock \bibinfo{journal}{\emph{Transactions of the Association for
  Computational Linguistics}}  \bibinfo{volume}{8} (\bibinfo{date}{04}
  \bibinfo{year}{2020}), \bibinfo{pages}{247--263}.
\newblock


\bibitem[Silva et~al\mbox{.}(2022)]%
        {Silva2020UsingCI}
\bibfield{author}{\bibinfo{person}{Andrew Silva}, \bibinfo{person}{Rohit
  Chopra}, {and} \bibinfo{person}{Matthew Gombolay}.}
  \bibinfo{year}{2022}\natexlab{}.
\newblock \showarticletitle{Cross-Loss Influence Functions to Explain Deep
  Network Representations}. In \bibinfo{booktitle}{\emph{Proceedings of The
  25th International Conference on Artificial Intelligence and Statistics}}
  \emph{(\bibinfo{series}{Proceedings of Machine Learning Research},
  Vol.~\bibinfo{volume}{151})}, \bibfield{editor}{\bibinfo{person}{Gustau
  Camps-Valls}, \bibinfo{person}{Francisco J.~R. Ruiz}, {and}
  \bibinfo{person}{Isabel Valera}} (Eds.). \bibinfo{publisher}{PMLR},
  \bibinfo{pages}{1--17}.
\newblock


\bibitem[Van~der Linden and Glas(2000)]%
        {CAT2000}
\bibfield{editor}{\bibinfo{person}{Wim~J Van~der Linden} {and}
  \bibinfo{person}{Cees~AW Glas}} (Eds.). \bibinfo{year}{2000}\natexlab{}.
\newblock \bibinfo{booktitle}{\emph{Computerized Adaptive Testing: Theory and
  Practice}}.
\newblock \bibinfo{publisher}{Springer Dordrecht}.
\newblock


\bibitem[Wang and Chang(2011)]%
        {WC11}
\bibfield{author}{\bibinfo{person}{Chun Wang} {and} \bibinfo{person}{Hua-Hua
  Chang}.} \bibinfo{year}{2011}\natexlab{}.
\newblock \showarticletitle{Item Selection in Multidimensional Computerized
  Adaptive Testing—Gaining Information from Different Angles}.
\newblock \bibinfo{journal}{\emph{Psychometrika}}  \bibinfo{volume}{76}
  (\bibinfo{year}{2011}), \bibinfo{pages}{363–384}.
\newblock


\bibitem[Wang et~al\mbox{.}(2011)]%
        {WCB11}
\bibfield{author}{\bibinfo{person}{Chun Wang}, \bibinfo{person}{Hua-Hua Chang},
  {and} \bibinfo{person}{Keith~A Boughton}.} \bibinfo{year}{2011}\natexlab{}.
\newblock \showarticletitle{Kullback–Leibler Information and Its Applications
  in Multi-Dimensional Adaptive Testing}.
\newblock \bibinfo{journal}{\emph{Psychometrika}}  \bibinfo{volume}{76}
  (\bibinfo{year}{2011}), \bibinfo{pages}{13--39}.
\newblock


\bibitem[Wang et~al\mbox{.}(2021)]%
        {wang2021combating}
\bibfield{author}{\bibinfo{person}{Xiaojie Wang}, \bibinfo{person}{Rui Zhang},
  \bibinfo{person}{Yu Sun}, {and} \bibinfo{person}{Jianzhong Qi}.}
  \bibinfo{year}{2021}\natexlab{}.
\newblock \showarticletitle{Combating Selection Biases in Recommender Systems
  with a Few Unbiased Ratings}. In \bibinfo{booktitle}{\emph{Proceedings of the
  14th ACM International Conference on Web Search and Data Mining}}
  \emph{(\bibinfo{series}{WSDM '21})}. \bibinfo{publisher}{Association for
  Computing Machinery}, \bibinfo{address}{New York, NY, USA},
  \bibinfo{pages}{427--435}.
\newblock


\bibitem[Wang et~al\mbox{.}(2020)]%
        {wang2020instructions}
\bibfield{author}{\bibinfo{person}{Zichao Wang}, \bibinfo{person}{Angus Lamb},
  \bibinfo{person}{Evgeny Saveliev}, \bibinfo{person}{Pashmina Cameron},
  \bibinfo{person}{Yordan Zaykov}, \bibinfo{person}{Jos{\'e}~Miguel
  Hern{\'a}ndez-Lobato}, \bibinfo{person}{Richard~E Turner},
  \bibinfo{person}{Richard~G Baraniuk}, \bibinfo{person}{Craig Barton},
  \bibinfo{person}{Simon~Peyton Jones}, {et~al\mbox{.}}}
  \bibinfo{year}{2020}\natexlab{}.
\newblock \showarticletitle{Instructions and guide for diagnostic questions:
  The neurips 2020 education challenge}.
\newblock \bibinfo{journal}{\emph{arXiv preprint arXiv:2007.12061}}
  (\bibinfo{year}{2020}).
\newblock


\bibitem[Weiss(1982)]%
        {Weiss82}
\bibfield{author}{\bibinfo{person}{David~J. Weiss}.}
  \bibinfo{year}{1982}\natexlab{}.
\newblock \showarticletitle{Improving Measurement Quality and Efficiency with
  Adaptive Testing}.
\newblock \bibinfo{journal}{\emph{Applied Psychological Measurement}}
  \bibinfo{volume}{6}, \bibinfo{number}{4} (\bibinfo{year}{1982}),
  \bibinfo{pages}{473--492}.
\newblock


\bibitem[Weiss and Kingsbury(1984)]%
        {WK1984}
\bibfield{author}{\bibinfo{person}{David~J. Weiss} {and}
  \bibinfo{person}{G.~Gage Kingsbury}.} \bibinfo{year}{1984}\natexlab{}.
\newblock \showarticletitle{Application of computerized adaptive testing to
  educational problems}.
\newblock \bibinfo{journal}{\emph{Journal of Educational Measurement}}
  \bibinfo{volume}{21}, \bibinfo{number}{4} (\bibinfo{year}{1984}),
  \bibinfo{pages}{361--375}.
\newblock


\bibitem[Yang et~al\mbox{.}(2022)]%
        {YANG2022100104}
\bibfield{author}{\bibinfo{person}{Albert~C.M. Yang}, \bibinfo{person}{Brendan
  Flanagan}, {and} \bibinfo{person}{Hiroaki Ogata}.}
  \bibinfo{year}{2022}\natexlab{}.
\newblock \showarticletitle{Adaptive formative assessment system based on
  computerized adaptive testing and the learning memory cycle for personalized
  learning}.
\newblock \bibinfo{journal}{\emph{Computers and Education: Artificial
  Intelligence}}  \bibinfo{volume}{3} (\bibinfo{year}{2022}),
  \bibinfo{pages}{100104}.
\newblock


\bibitem[Yao(2013)]%
        {Yao13}
\bibfield{author}{\bibinfo{person}{L. Yao}.} \bibinfo{year}{2013}\natexlab{}.
\newblock \showarticletitle{Comparing the Performance of Five Multidimensional
  CAT Selection Procedures With Different Stopping Rules}.
\newblock \bibinfo{journal}{\emph{Applied Psychological Measurement}}
  \bibinfo{volume}{37}, \bibinfo{number}{1} (\bibinfo{year}{2013}),
  \bibinfo{pages}{3--23}.
\newblock


\bibitem[Yu et~al\mbox{.}(2020)]%
        {yu2020influence}
\bibfield{author}{\bibinfo{person}{Jiangxing Yu}, \bibinfo{person}{Hong Zhu},
  \bibinfo{person}{Chih-Yao Chang}, \bibinfo{person}{Xinhua Feng},
  \bibinfo{person}{Bowen Yuan}, \bibinfo{person}{Xiuqiang He}, {and}
  \bibinfo{person}{Zhenhua Dong}.} \bibinfo{year}{2020}\natexlab{}.
\newblock \showarticletitle{Influence function for unbiased recommendation}. In
  \bibinfo{booktitle}{\emph{Proceedings of the 43rd International ACM SIGIR
  Conference on Research and Development in Information Retrieval}}.
  \bibinfo{pages}{1929--1932}.
\newblock


\bibitem[Zhuang et~al\mbox{.}(2022a)]%
        {rcat}
\bibfield{author}{\bibinfo{person}{Yan Zhuang}, \bibinfo{person}{Qi Liu},
  \bibinfo{person}{Zhenya Huang}, \bibinfo{person}{Zhi Li},
  \bibinfo{person}{Binbin Jin}, \bibinfo{person}{Haoyang Bi},
  \bibinfo{person}{Enhong Chen}, {and} \bibinfo{person}{Shijin Wang}.}
  \bibinfo{year}{2022}\natexlab{a}.
\newblock \showarticletitle{A Robust Computerized Adaptive Testing Approach in
  Educational Question Retrieval}. In \bibinfo{booktitle}{\emph{Proceedings of
  the 45th International ACM SIGIR Conference on Research and Development in
  Information Retrieval}}. \bibinfo{pages}{416–426}.
\newblock
\showISBNx{9781450387323}


\bibitem[Zhuang et~al\mbox{.}(2022b)]%
        {zhuang2022fully}
\bibfield{author}{\bibinfo{person}{Yan Zhuang}, \bibinfo{person}{Qi Liu},
  \bibinfo{person}{Zhenya Huang}, \bibinfo{person}{Zhi Li},
  \bibinfo{person}{Shuanghong Shen}, {and} \bibinfo{person}{Haiping Ma}.}
  \bibinfo{year}{2022}\natexlab{b}.
\newblock \showarticletitle{Fully Adaptive Framework: Neural Computerized
  Adaptive Testing for Online Education}. In
  \bibinfo{booktitle}{\emph{Proceedings of the AAAI Conference on Artificial
  Intelligence}}, Vol.~\bibinfo{volume}{36}. \bibinfo{pages}{4734--4742}.
\newblock


\end{thebibliography}

\end{document}